\documentclass[11pt,a4paper]{article}
\usepackage[hyperref]{acl2017}
\usepackage{times}
\usepackage{latexsym}

\usepackage{url}

\usepackage{tabularx}
\usepackage{subcaption}
\usepackage{tabularx}
\usepackage{booktabs}
\usepackage{multirow}
\usepackage{graphicx}
\usepackage{amsmath}
\usepackage{comment}
\usepackage{color}
\usepackage{xcolor,colortbl}
\usepackage{enumitem}

\aclfinalcopy 

\title{Modelling Protagonist Goals and Desires in First-Person Narrative}

\author{Elahe Rahimtoroghi, Jiaqi Wu, Ruimin Wang, \\ 
\textbf{Pranav Anand} and \textbf{Marilyn A Walker} \\
 University of California Santa Cruz \\
  Santa Cruz, CA, US \\
  {\tt \{erahimto,jwu64,ruiwang,panand,mawalker\}@ucsc.edu}}

\date{}

\begin{document}
\maketitle
\begin{abstract}
Many genres of natural language text are narratively structured, a
testament to our predilection for organizing our experiences as
narratives. There is broad consensus that understanding a narrative
requires identifying and tracking the goals and desires of the
characters and their narrative outcomes.  However, to date, there has been
limited work on computational models for this problem.  We introduce a
new dataset,  \textbf{DesireDB}, which includes gold-standard
labels for identifying statements of desire, textual evidence for
desire fulfillment, and annotations for whether the stated desire is
fulfilled given the evidence in the narrative context.  We 
report experiments on tracking desire fulfillment using different methods, and show
that LSTM Skip-Thought model achieves F-measure of 0.7 on our corpus. 
\end{abstract}

\section{Introduction}
\label{intro}

Humans appear to organize and remember everyday experiences by
imposing a narrative structure on them
\cite{nelson89,ThorneNam09,bruner91,Mcadamsetal06}, 
and many genres of natural language text are therefore narratively
structured, e.g. dinner table conversations, news articles, user
reviews and blog posts \cite{Polanyi89,Jurafskyetal14,bell2005news,Gordonetal11}.
Moreover, there is broad consensus that understanding a narrative
involves activating a representation, early in the narrative, of the
protagonist and  her goals and desires, and then maintaining
that representation as the narrative evolves, as a vehicle for
explaining the protagonist's actions and tracking narrative outcomes
\cite{Elson-Thesis12,RappGerrig06,trabassoVandenBroek85,Lehnert81}.

To date, there has been limited work on computational models for
recognizing the expression of the protagonist's goals and desires in
narrative texts, and tracking their corresponding narrative
outcomes. We introduce a new corpus \textbf{DesireDB} of $\sim$3,500
first-person informal narratives with annotations for desires and
their fulfillment status, available online.\footnote{https://nlds.soe.ucsc.edu/DesireDB}
Because first-person narratives often
revolve around the narrator's private states and goals \cite{Labov72},
this corpus is highly suitable as a testbed for identifying human
desires and their outcomes. Moreover, first-person narratives allow
the narrative protagonist (first-person) to be easily identified and
tracked.  Figure~\ref{fig:desire-example} illustrates examples of
desire and goal expressions in our corpus.

\begin{figure}[t]
\small
\begin{tabularx}{\columnwidth}{X}
\toprule
People did seem pleased to see me but all I \textbf{[wanted to]} do was talk to a particular friend.\\
\midrule 
I'm off this weekend and had really \textbf{[hoped to]} get out and dance.\\
\midrule
We \textbf{[decided to]} just go for a walk and look at all the sunflowers in the neighborhood. \\
\midrule
I \textbf{[couldn't wait to]} get out of our cheap and somewhat charming hotel and show James a little bit of Paris.\\
\midrule
We drove for just over an hour and \textbf{[aimed to]} get to Trinity beach to set up for the night. \\
\midrule
She called the pastor, and he had time, too, so, we \textbf{[arranged to]} meet Saturday at 9am.\\
\midrule
Even though my deadline wasn't until 4 p.m., I \textbf{[needed to]} write the story as quickly as possible.\\
\bottomrule
\end{tabularx}
\caption{Desire expressions in personal narratives}
\label{fig:desire-example}
\end{figure}

DesireDB is open domain. It contains a broad range of expressions of
desires and goal statements in personal narratives. It also
includes the narrative context for each desire statement as shown in
Figure~\ref{fig:mt-example}.  We include both prior and post context
of the desire expressions, since theories of narrative structure
suggest that the evaluation points of a narrative can precede the
expression of the events, goals and desires of the
narrator~\cite{Labov72,Swansonetal14b}.

Our approach builds on seminal work on a computational model of
Lehnert's plot units, that applied modern NLP tools to tracking
narrative affect states in Aesop's Fables
\cite{Goyaletal10,Lehnert81,goyal2013computational}. Our framing of
the problem is also inspired by recent work that identifies three
forms of desire expressions in short narratives from MCTest and
SimpleWiki and develops models to predict whether desires are
fulfilled or unfulfilled \cite{Chaturvedietal16}. However DesireDB's
narrative and sentence structure is more complex than
either MCTest or SimpleWiki
\cite{richardson2013mctest,coster2011simple}.

We propose new features (Sec~\ref{sub:features}), as
well as testing features used in previous work, and apply different
classifiers to model desire fulfillment in our corpus. We also
directly compare to results on  MCTest and SimpleWiki
(Sec~\ref{sub:comparison}). We apply LSTM models that distinguish
between prior and post context and capture the flow of the
narrative. Our best system, a Skip-Thought RNN model, achieves an F-measure of
0.70, while a logistic regression system achieves 0.66.
Our models and features outperform
\newcite{Chaturvedietal16} on MCTest and SimpleWiki, while providing
new results for a new corpus for tracking desires in first-person narratives.
Moreover, analysis of our results shows that features representing
the  discourse
structure (such as overt discourse relation markers)
are the best predictors of fulfillment status of a desire or goal. 
We also show that both prior and post
context are important for this task.  

We discuss related work in Sec.~\ref{related} and describe our corpus and annotations in Sec.~\ref{dataset}. Section~\ref{model} presents our features and methods for modeling desire fulfillment in narratives along with the experiments and results including comparison to previous work. Finally, we present conclusions and future directions in Sec.~\ref{conclusion}.

\begin{figure}[t]
\small
\begin{tabularx}{\columnwidth}{X}
\toprule
\textbf{Prior-Context}: 
(1) I ran the Nike+ human Race 10K new York in under 57 minutes!
(2) Then at the all-American rejects concert, I somehow ended up right next to this really cute guy and he seemed interested in me. 
(3) Was I imagining things? He was really nice; 
(4) I dropped something and it was dark, he bent with his cell phone light to help me look for it. 
(5) We spoke a little, but it was loud and not suited for conversation there.

\textbf{Desire-Expression-Sentence}: I \textbf{[had hoped to]} ask him to join me for a drink or something after the show (if my courage would allow such a thing) but he left before the end and I didn't see him after that.

\textbf{Post-Context}: 
(1) Maybe I'll try missed connections lol.
(2) I didn't want to tell him I think he's cute or make any gay references during the show because if I was wrong that would make standing there the whole rest of the concert too awkward...
(3) Afterward, I wandered through the city making stops at several bars and clubs, met some new people, some old people
(4) As in people I knew - I actually didn't met any old people, unless you count the tourist family whose dad asked me about my t-shirt.
(5) And when I thought the night was over (and the doorman of the club did insist it was over) I met this great guy going into the subway.
\\ \bottomrule
\end{tabularx}
\caption{A desire expression with its surrounding context extracted from a personal narrative}
\label{fig:mt-example}
\end{figure}

\section{Related Work}
\label{related}

There has recently been an upsurge in interest in 
computational models of narrative structure~\cite{Lehnert81,wilensky82} and story understanding~\cite{Rahimtoroghietal16,Swansonetal14b,OuyangMcKeown15,OuyangMcKeown14}.
However
there has been limited work on computational models for
recognizing the expression of the protagonist's goals and desires in
narrative genres. 

Our approach builds on work by \newcite{goyal2013computational} that
applied modern NLP tools to track narrative affect states in Aesop's
Fables \cite{Goyaletal10}.  They present a system called AESOP that
uses a number of existing resources to identify affect states of the
characters as part of deriving plot units. The motivation of modeling
plot units is the idea that emotional reactions are central to the
notion of a narrative and the main plot of a story can be modeled by
tracking the transition between the affect states~\cite{Lehnert81}.
The AESOP system identifies affect states and creates links between
them to model plot units and is evaluated on a small set of
two-character fables.  They performed a manual annotation to examine
different types of affect expressions in the narratives.  Their study
shows that many affect states arise from events where a character is
acted upon in positive or negative ways, not explicit expression of
emotions.  They also show that most of the affect states emerge by the
expression of goals and plans and goal completion. Some of our
features are motivated by the idea that implicit sentiment polarity
can represent success or failure of goals and can be used to better
model desire and goal fulfillment in a narrative \cite{Reedetal17},
although we cannot directly compare our findings to theirs because
their annotations are not publicly available.

\newcite{Chaturvedietal16} exploit two deliberately simplified
datasets in order to model desire and its fulfillment: \textbf{MCTest}
which contains 660 stories limited to content understandable by 7-year
old children, and, \textbf{SimpleWiki} created from a dump of the
Simple English Wikipedia discarding all the lists, tables and
titles. They use desire statements matching a list of three verb
phrases, \textit{wanted to, hoped to,} and \textit{wished to}. Their
context representation consists of five or fewer sentences following
the desire expression. They use BOW (Bag of Words) as baseline and apply unstructured
and structured models for desire fulfillment modeling with different
features motivated by narrative structure.  Their best result is
achieved with a structured prediction model called Latent Structured
Narrative Model (LSNM) which models the evolution of the narrative by
associating a latent variable with each fragment of the context in the
data. Their best unstructured model is a Logistic Regression
classifier that uses all of their features.

Recent work on computational models of semantics provides an evaluation test for story understanding~\cite{mostafazadeh2017lsdsem}. The task includes four-sentence stories, each with two possible endings where only one is correct. The goal is for each system to select the correct ending of the story by modeling different levels of semantics in narratives, such as  lexical, sentential and discourse-level. The highest performing model with 75\% accuracy used a linear regression classifier with several features such as neural language models and stylistic features to model the story coherence~\cite{schwartz2017story}. The results from other systems showed that sentiment is an important factor and using only sentiment features could achieve about 65\% accuracy on the test.

\section{DesireDB Corpus}
\label{dataset}

DesireDB aims to provide a testbed for modeling desire and goals in
personal narrative and predicting their fulfillment status.  We develop
a systematic method to identify desire and goal statements, and then
collect annotations to create gold-standard labels of fulfillment status
as well as spans of text marked as evidence.

\subsection{Identifying Desires and Goals}
\label{desire-corpus}

Our corpus is a subset of the Spinn3r corpus
\cite{icwsm11,Burtonetal09}, consisting of first-person narratives
from six personal blog domains: \emph{livejournal.com, wordpress.com,
  blogspot.com, spaces.live.com, typepad.com, travelpod.com}. To create our dataset,
  we select only desire expressions involving some version
of the first-person. In first-person narratives, the narrator and protagonist
naturally align which makes it much easier to identify and track the protagonist than in fiction or historical genre. Thus, selecting narrative
passages with expressions of desire relating to the first-person
are very likely to discuss subsequent behaviors to achieve that desire
and the end result. Put simply, zooming in on first-person desires means that desire
and its aftermath are more likely to be highly topical for the narrative.
This corpus, then, is highly suitable as a testbed for modeling human desires and their
fulfillment.

Human desires and goals can be expressed linguistically in many different ways,
including both explicit verbal and nominal markers of desire or necessity 
(e.g., \emph{want, hope}) and more general markers of urges (e.g., \emph{craving, hunger, thirst}).  To systematically discover
predicates that specify desires,  we browsed 
FrameNet 1.7 ~\cite{Bakeretal98} selecting frames that seemed likely to contain lexical
units specifying desires: \textit{Being-necessary, Desiring,
  Have-as-a-demand, Needing, Offer, Purpose, Request, Required-event,
  Scheduling, Seeking, Seeking-to-achieve, Stimulus-focus,
  Stimulate-emotion,} and \textit{Worry}. We then
selected 100
representative instances of that frame in English Gigaword \cite{parker2011english} by first
selecting the 10 most frequent lexical units in that frame,
and then selecting 10 random instances per lexical unit.  One of the
authors examined each set of 100 instances, estimating for each
sentence whether the predicate specifies a goal that the surrounding
text picks up on.
Because we were looking for predicates that reliably specify desires that motivate a protagonist's actions, we eliminated frames where less than 80\% of the sentences showed this characteristic. 

This resulted in a downsample to the following four frames: \textit{Desiring, Needing, Purpose,} and \textit{Request}. 
We selected only the verbal lexical units
because we found that verbs were more likely to introduce goals than
nouns or adjectives. We examined 100 instances for each verbal lexical
unit, discarding as before. This resulted in
37 verbs.  For each verb, we systematically constructed and coded
all past forms of the verb (e.g., \emph{was [verb]ing}, \emph{had
  [verb]ed}, \emph{had been [verb]ing}, \emph{[verb]ed}, \emph{didn't
  [verb]}, etc.) because we posited that morphological form itself may
convey likelihood of fulfillment (e.g., a past perfect \emph{I had
  wanted to ...} signals that something changed, either the
desire or fulfillment).  We initially experimented with both past and
(historical) present, but 
past tense verb patterns resulted in much 
higher precision. We counted the instances of these patterns
in our dataset, and retained only those lemmas with at least 1000
instances across the corpus.

We extract stories containing the verbal patterns of desire,
with five sentences before and after the desire expression sentence as
context (See Fig.~\ref{fig:mt-example}). Our annotation results provide support that the evidence of
desire fulfillment can be expressed before the desire statement. We
also study the effect of prior and post context in understanding
desire fulfillment in our experiments (Section~\ref{model}) and show
that using the narrative context preceding the desire statement 
improves the results.

\begin{figure}[t]
\small
\begin{tabularx}{\columnwidth}{X}
\toprule
\textbf{Data-Instance:} 

Prior-Context:
ConnectiCon!!! Ya baby, we did go this year as planned! Though this year we weren't in the artist colony, so I didn't see much point in posting about it before hand.

Desire-Expression-Sentence:
This year we [wanted to] be part of the main crowd.

Post-Context:
We wanted to get in on all the events and panels that you cant attend when watching over a table. And this year we wanted to cosplay! My hubby and I decided to dress up like aperture Science test subjects from the PC game portal. It was a good and original choice, as we both ended up being the only portal related people in the con (unless there were others who came late in the evening we didn't see) It was loads of fun and we got a surprising amount of attention. \\
\midrule
\textbf{Annotations:}

Fulfillment-Label:
Fulfilled

Fulfillment-Agreement-Score: 3

Evidence:
Though this year we weren't in the artist colony. We wanted to get in on all the events and panels that you cant attend when watching over a table.

Evidence-Overlap-Score: 3 \\
\bottomrule
\end{tabularx}
\caption{Example of data in DesireDB}
\label{fig:annotated}
\end{figure}







\begin{table}[t]
\footnotesize
\centering
\begin{tabularx}{3in}{p{0.65in} c c c c c}
\toprule
\bf Pattern & \bf Count & \bf Ful & \bf Unf & \bf Unk & \bf None \\
\midrule
wanted to & 2,510 & 49\% &	35\%	& 14\%	& 2\% \\
needed to & 202 & 65\%	& 16\%	& 16\% & 	3\% \\
ordered & 201 & 71\% &	21\% &	6\% & 	2\% \\
arranged to & 199 & 68\% &	13\%	& 16\%	& 3\%\\
decided to & 68 & 87\% & 	9\% & 	4\% & 	0\% \\
hoped to & 68 & 19\% & 	68\% & 	12\%	& 1\% \\
couldn't wait & 68 & 79\% &	3\% &	15\%	& 3\% \\
wished to & 66 & 27\% &	35\% &	30\%	& 8\% \\
scheduled & 60 & 43\% & 	25\% & 	27\%	& 5\% \\
asked for & 60 & 53\%	& 27\% &	15\%	& 5\% \\
required & 58 & 69\%	& 16\%	& 15\% & 	0\% \\
requested & 30 & 60\% & 	20\% &	20\% &	0\% \\
demanded & 30 & 60\% & 	23\% &	17\% &	0\% \\
ached to & 20 & 50\%	& 40\%  & 	10\% & 	0\% \\
aimed to & 20 & 55\%	& 30\%	& 15\%  &	 0\% \\
desired to & 20 & 50\%	& 25\% & 	25\% & 	0\% \\
\midrule
Total & 3,680 & 53\%  & 31\% & 14\% & 2\% \\
\bottomrule
\end{tabularx}
\caption{\label{tab:corpus-labels} Distribution of desire verbal patterns and fulfillment labels in DesireDB}
\end{table}

\subsection{Data Annotation}
\label{gold-standard}

We extracted $\sim$600K desire expressions with their
context, and then sample 3,680 instances for annotation. This subset
consists of 16 verbal patterns (when collapsing all morphological
forms to their head word).  A group of pre-qualified Mechanical
Turkers then labelled each instance. The annotators labelled
the fulfillment status of the desire expression sentence
based on the prior and post context, by choosing from three labels:
\emph{Fulfilled, Unfulfilled,} and \textit{Unknown from the
  context}. They were also asked to mark the evidence for the label
they had chosen by specifying a span of text in the narrative.  
For each data instance, we asked the Turkers to mark the subject of the desire expression and determine if the expressed desire is hypothetical (e.g., a conditional sentence) or not.

The annotators were selected from a list of pre-qualified workers who had
successfully passed a test on a textual entailment task with 100\%
correct answers. They were provided with detailed instructions and
examples as to how to label the desires and mark the evidence. We also
specified the desire expression verbal pattern using square brackets
(as shown in Fig.~\ref{fig:desire-example} and~\ref{fig:mt-example})
for more clarity. Three annotators were assigned to work on each data
instance. To generate the gold-standard labels we used majority vote
and the cases with no agreement were labeled as \emph{`None'}.

Table~\ref{tab:corpus-labels} reports the distribution of data and
gold-standard labels (Ful:Fulfilled, Unf:Unfulfilled, Unk:Unknown from
the context). About half of the desire expressions (53\%) were labeled
\textit{Fulfilled} and about one third (31\%) were labeled
\textit{Unfulfilled}. The annotators didn't agree on about 2\% of the
instances, that were labeled \textit{None}. As
Tabel~\ref{tab:corpus-labels} shows, the distribution of labels is not
uniform across different verbal patterns. For instance,
\textit{decided to} and \textit{couldn't wait} are highly skewed
towards Fulfilled as opposed to \textit{hoped to} which includes 68\%
Unfulfilled instances. Some patterns seem to be harder to annotate,
like \textit{wished to}, which has the highest rate of Unknown (30\%)
and None (8\%) among all.

Other than fulfillment status, for each data instance in our corpus we
include the agreement-score which is the number of annotators that
agreed on the assigned label.  In addition, we provide the
\textit{evidence} as a part of the DesireDB data, by merging the text
spans marked by the annotators as evidence. We compared the evidence
spans pairwise to measure the overlap-score, indicating the number of
pairs of annotators with overlapping responses.  An example
is shown in Figure~\ref{fig:annotated}. The first
part is the extracted data including the desire expression with prior and post context, and the second part is the gold-standard annotations.

To assess inter-annotator agreement for Fulfillment, we
calculated Krippendorff-alpha Kappa~\cite{krippendorff1970bivariate,krippendorff2004content} for pairwise inter-annotator
reliability, and, the average of Kappa between each annotator and the
majority vote. These two metrics are 0.63 and 0.88 respectively.
Overall, 66\% of the data was labeled with total agreement (where all
three annotators agreed on the same label) and about 32\% of data was
labeled by two agreements and one disagreement. We also examined the
agreements across each label separately. For \textit{Fulfilled} class,
total agreement rate is 75\%, which for \textit{Unfulfilled} is 67\%,
and on \textit{Unknown from the context} is 41\%. We believe this
indicates that annotating unfulfilled desires was harder than
fulfilled cases.  For evidence marking, in 79\% of the data all three
annotators marked overlapping spans. 



\section{Modeling Desire Fulfillment}
\label{model}

We conducted a range of experiments on 
predicting fulfillment status of desires and goals, using different features and models, including LSTM architectures that can encode the sequential structure of the narratives. We first describe
our features and models. Then, we present our feature analysis study
to examine their importance in modeling
fulfillment. Finally we provide results of
direct comparison to previous work on the existing corpora.

\begin{figure}[t]
\small
\begin{tabularx}{\columnwidth}{X}
\toprule
Sentiment: Negative

Prior-Context(4):
"I had been working for hours on boring paperwork and financial stuff, and I was really crabby."\\
\midrule
Sentiment: Negative

Prior-Context(5):
I decided it was time to take a break and thought, should I read a magazine or watch best Week Ever?\\
\midrule
Sentiment: Negative

Desire-Epxression-Sentence:
But I realized that what I really \textbf{[wanted to]} do was go for a run!\\
\midrule
Sentiment: Positive

Post-Context(1):
That was pretty amazing, to transition mentally from 'having to' to 'wanting to' run.\\
\midrule
Sentiment: Positive

Post-Context(2):
So I did a quick, fun 2.75 miles.\\
\bottomrule
\end{tabularx}
\caption{Example of sentiment features, where prior context is negative while the post context is positive, implying fulfillment of the desire}
\label{fig:sentiment-example}
\end{figure}


\subsection{Features Description}
\label{sub:features}

In our original informal examination of the DesireDB development data, we noticed
several ways that a writer can signal (lack of) fulfillment of a desire like 
``I hoped to pick up a dictionary''. First, they may mention an outcome that entails
 (``The book I bought was...'') or strongly implies fulfillment (``I went back home happily.'').
 However, we noticed that in many cases of fulfillment, the `marker' was simply the absence of
 any mention that things went wrong. For lack of fulfillment, while we found cases where writers 
 explicitly state that their desire wasn't met, we noted many instances where evidence came from
 mentioning that an enabling condition for fulfillment wasn't met (``The bookstore was closed.'').
 
 True machine understanding of these kinds of narrative structures requires robust models of the complex interplay of
 semantics (including negation) as well as world knowledge about the scripts for tasks like buying books,
 including what count as enabling conditions and entailers for fulfillment. While we hope to explore
 more articulated models in the future, for our experiments we considered reasonable proxies for the conditions
 mentioned above using existing resources (note that we also tested LSTM models described below, which may implicitly learn
 such relationships with sufficient data). One set (\textbf{Desire Features}) indexes properties of the
 desire expression (e.g., the desire verb) as well as overlap between the desired object/event and the surrounding
 context. The remaining features attempt to
 find general markers for success or failure. One set (\textbf{Discourse Features}) looks for overt discourse
 relation markers that signal violation of expectation (e.g., `but', `however') or its opposite (e.g., `so').
 Another uses the Connotation Lexicon \cite{Fengetal13} to model whether the context provides a positive or negative
 event. All of these features are inspired by~\newcite{Chaturvedietal16}. Finally, motivated by the AESOP modeling of affect states for identifying plot units~\cite{goyal2013computational}, one set of features
  (\textbf{Sentiment-Flow-Features}) indexes whether there has been a change in sentiment in the surrounding
  context (which might be the mention of a thwarted effort or a hard won victory). Figure~\ref{fig:sentiment-example}
  provides an example of this.


In addition to a BOW (Bag of Words) baseline, we extracted the four types of features mentioned above. For
features that examine the context around the desire expression, our experiments
used the pre-context, the post-context, or both, as discussed below; context features are computed per
sentence $i$ of the context. 
We also tested various ablations
of these features described below as well.  We now describe the full set of features in more detail.

\noindent \textbf{Desire-Features.} From a desire expression of the form `X Ved S', we extract
the lexical feature \emph{Desire-Verb}, the lemma for V. We also extract a list of \emph{focal words}, 
the content words in embedded sentence S. In Figure~\ref{fig:sentiment-example}, these
are `do', `go', and `run'. The features \emph{Focal-\{Word,Synonym,Antonym\}-Mention-$i$} 
counts how many times each word, its synonyms, or its antonyms in WordNet~\cite{WordNet} are in the context, respectively. Similarly,
\emph{Desire-Subject-Mention-$i$} marks if subject X is mentioned in the context. Finally, boolean
\emph{First-Person-Subject} indicates if X is first person (`I', `we').

\noindent \textbf{Discourse-Features.} This class of features count how many of two classes of discourse
relation markers (\emph{Violated-Expectation--$i$} vs. \emph{Meeting-Expectation--$i$}) occur in the context. 
For the classes, we manually coded all overt discourse relation markers in the 
Penn Discourse Treebank three ways(violation, meeting, or neutral), 
leading to 15 meeting markers (`accordingly', `so', `ultimately', `finally') and 31 violating (`although', `rather', `yet', `but'). In addition, we also tracked the presence of the most frequent of these (`so' and `but', respectively)
in the desire sentence itself by the booleans \emph{So-Present} and \emph{But-Present}.
 
\noindent \textbf{Connotation-Features.}  Beyond the use of WordNet expansion for \emph{Focal-Word-Mention-$i$},
we also used the Connotation Lexicon \cite{Fengetal13}, a lexical resource marking very 
general connotation polarities (positive or negative) of words (as opposed to more specific 
sentiment lexicons). \emph{Connotation-Agree-$i$} counts for each word $w$ in \emph{focal words} the number of words in the context that have the same connotation polarity as $w$.
\emph{Connotation-Disgree-$i$} is defined similarly.

\noindent \textbf{Sentiment-Flow-Features.} To model affect states, we
compute a sentiment score for the desire expression sentence as well
as each sentence in the context. Then for each sentence of the
context, the booleans \emph{Sentiment-Agree-$i$} and
\emph{Sentiment-Disagree-$i$} mark whether that sentence and the
desire expression sentence have the same sentiment polarity (see
Figure~\ref{fig:sentiment-example}).  While there is evidence
suggesting that models of implicit sentiment (e.g.,
\cite{Goyaletal10,Reedetal17}) could do much better at tracking affect
states, here we use the Stanford Sentiment system~\cite{Socheretal13}.

%

\begin{table}[t]
\footnotesize
\centering
\begin{tabularx}{3in}{c c c c c}
\toprule
\bf Fulfilled & \bf Unfulfilled & \bf Unknown & \bf None  & \bf Total\\
\midrule
1,366 & 953 & 380 & 70 & 2,780 \\
\bottomrule
\end{tabularx}
\caption{\label{tab:corpus-subset} Simple-DesireDB dataset}
\end{table}

\begin{table*}[t]
\footnotesize
\centering
\begin{tabularx}{6in}{p{0.7in}| p{0.45in} | c c c | c c c | c c c}
\toprule
\bf Method & \bf Features  & \bf Ful-P & \bf Ful-R & \bf Ful-F1 & \bf Unf-P & \bf Unf-R & \bf Unf-F1  & \bf Precision & \bf Recall & \bf F1\\
\toprule
Skip-Thought & BOW & 0.75 & 0.70 & 0.72 & 0.54 & 0.61 & 0.57 & 0.65 & 0.65 & 0.65 \\
& \bf ALL & \bf 0.80 & \bf 0.71 & \bf 0.75 & \bf 0.59 & \bf 0.70 & \bf 0.64 & \bf 0.70 & \bf 0.70 & \bf 0.70 \\
\midrule
CNN-RNN  & BOW & 0.75	 & 0.73 & 0.74 & 0.57 & 0.60 & 0.58 & 0.66 & 0.66 & 0.66\\
 & ALL & 0.75 & 0.79 & 0.77 & 0.61 & 0.56 & 0.59 & 0.68 & 0.68 & 0.68 \\
\bottomrule
\end{tabularx}
\caption{\label{tab:NN-Features} Results of LSTM models on Simple-DesireDB}
\end{table*}

\begin{table*}[t]
\footnotesize
\centering
\begin{tabularx}{5.6in}{p{0.9in} | c c c | c c c | c c c}
\toprule
\bf Data & \bf Ful-P & \bf Ful-R & \bf Ful-F1 & \bf Unf-P & \bf Unf-R & \bf Unf-F1  & \bf Precision & \bf Recall & \bf F1\\
\midrule
Desire & 0.74 & 0.75 & 0.75 & 0.57 & 0.56 & 0.57 & 0.66 & 0.66 & 0.66 \\
Desire+Prior & 0.78 & 0.73 & 0.75 & 0.58 & 0.65 & 0.61 & 0.68 & 0.69 & 0.68 \\
Desire+Post & 0.76 & 0.70 & 0.73 & 0.55 & 0.62 & 0.59 & 0.66 & 0.66 & 0.66 \\
\bf Desire+Context & \bf 0.80 & \bf 0.71 & \bf 0.75 & \bf 0.59 & \bf 0.70 & \bf 0.64 & \bf 0.70 & \bf 0.70 & \bf 0.70 \\
\bottomrule
\end{tabularx}
\caption{\label{tab:RNN-Skip} Results of Skip-Thought using different parts of data, with ALL features on Simple-DesireDB}
\end{table*}

\begin{table*}[t]
\footnotesize
\centering
\begin{tabularx}{6in}{p{0.6in}| p{0.5in} | c c c | c c c | c c c}
\toprule
\bf Method & \bf Features  & \bf Ful-P & \bf Ful-R & \bf Ful-F1 & \bf Unf-P & \bf Unf-R & \bf Unf-F1  & \bf Precision & \bf Recall & \bf F1\\
\toprule
Skip- & BOW & 0.78 & 0.78 & 0.78 & 0.57 & 0.56 & 0.57 & 0.67 & 0.67 & 0.67 \\
Thought & All & 0.78 & 0.79 & 0.79 & 0.58 & 0.56 & 0.57 & 0.68 & 0.68 & 0.68 \\
 & \bf Discourse  & \bf 0.80 & \bf 0.79 & \bf 0.80 & \bf 0.60 & \bf 0.60 & \bf 0.60 & \bf 0.70 & \bf 0.70 & \bf 0.70 \\
\midrule
Logistic  & BOW & 0.69 & 0.65 & 0.67 & 0.53 & 0.57 & 0.55 & 0.61 & 0.61 & 0.61 \\
Regression & All & 0.79 & 0.70 & 0.74 & 0.52 & 0.64 & 0.58 & 0.66 & 0.67 & 0.66 \\
 & Discourse  & 0.75 & 0.84 & 0.80 & 0.60 & 0.45 & 0.52 & 0.67 & 0.65 & 0.66 \\
\bottomrule
\end{tabularx}
\caption{\label{tab:final-test} Results of best LSTM model with different feature sets, compared to LR on DesireDB}
\end{table*}

\subsection{LSTM Models}
\label{sub:nn}
Our features are motivated by narrative
characteristics but do not directly capture the sequential structure of the
narratives. We thus apply neural network models
suitable for sequence learning, in order to directly encode
the order of the sentences in the story and distinguish
between prior and post context.
We use two different architectures of  LSTM (Long Short-Term Memory)~\cite{hochreiter1997long} models to generate sentence embeddings and then apply a three-layer RNN (Recurrent Neural Network) for classification.
We used Keras~\cite{chollet2015keras} as a deep learning toolkit for implementing our experiments.

\noindent \textbf{Skip-Thoughts}.
This is a sequential model that uses pre-trained skip-thoughts model~\cite{kiros2015skip} as the embedding of sentences. 
It first concatenates features, if any, with embeddings, and then uses LSTM to generate a single representation for the context sequence, which is the output of the last unit. That single representation is then concatenated with embedding-feature concatenation of desire sentence and is fed into a multi-layer network to yield a single binary output.




\noindent \textbf{CNN-RNN.}  The only difference between the CNN-RNN
model and Skip-Thought is that it uses the 1-dimensional convolution
with max-over-time pooling introduced in~\cite{kim2014convolutional}
to generate the sentence embedding from word embedding, instead of
using skip-thoughts. We use Google News Vectors
\cite{mikolov2013distributed} for the word embedding with
different sizes from 1 to 7 for the kernel.




For our experiments, we first constructed a subset of DesireDB that we will call {\bf Simple-DesireDB},
in order to be able to compare more directly to the models and data used 
in previous work. \newcite{Chaturvedietal16} used three verb
phrases to identify desire expressions (\textit{wanted to},
\textit{hoped to}, and \textit{wished to}), so we selected a portion
of our corpus including these patterns along with two other
expressions (\textit{couldn't wait to} and \textit{decided to}) to
have sufficient data for experiments. Table~\ref{tab:corpus-subset}
shows the distribution of labels in this subset. For classification
experiments we use data labeled as \textit{Fulfilled} and
\textit{Unfulfilled}, thus the majority class accuracy is 59\%. We split the data into Train (1,656), Dev (327), and
Test (336) sets for the experiments.

Results of our two LSTM models for Fulfilled (Ful) and Unfulfilled
(Unf) classes and the overall classification task (P:precision,
R:recall) on Simple-DesireDB are presented in
Table~\ref{tab:NN-Features}. ALL feature set includes all the features
described in Sec. 4.1 (without BOW). The results indicate that our
features can considerably improve the model, compared to the BOW baseline
(F1 improved from 0.65 to 0.70 for Skip-Thought).  We also conducted 4
sets of experiments to study the importance of prior, post and the
whole context in predicting fulfillment status, using our best model.
The results of Skip-Thought using different contextual representations
are in Table~\ref{tab:RNN-Skip} with ALL features. The results
indicate that adding features from prior context {\bf alone} improves the results.
The best
results are obtained by including the whole context and desire
sentence.

We then experimented with our best model on all of DesireDB. We also
trained Naive Bayes, SVM and Logistic Regression (LR) classifiers as
baselines, with the best results on the Dev set  achieved by Logistic
Regression. Table~\ref{tab:final-test} shows the results of
Skip-Thought and LR on DesireDB for different features on the test set. Our
feature ablation study on the Dev set, discussed in
Sec.~\ref{sub:feature-selection}, indicates that Discourse features
are better predictors of fulfillment status, so we present  results
using only Discourse features in addition to BOW and ALL.

All of the results indicate that similar features and methods achieve
better results for the \textit{Fulfilled} class as compared to
\textit{Unfulfilled}. We believe the reason is that identifying
unfulfillment of a desire or goal is a more difficult task, as
discussed in the annotation description in
Section~\ref{gold-standard}. To further our analysis on the annotation
disagreements, we examined the cases where only two
annotators agreed on the assigned label. From the expressions labeled
\textbf{Fulfilled} by two annotators, 64\% were labeled
\textit{Unknown from the context} by the disagreeing annotator, and
only 36\% were labeled \textit{Unfulfilled}. However, these numbers
for the \textbf{Unfulfilled} class are respectively 49\% and 51\%,
indicating a stronger disagreement between annotators when labeling
Unfulfilled expressions.

\subsection{Feature Selection Experiments}
\label{sub:feature-selection}



We used the InfoGain measure to rank features based on their
importance in modeling desire fulfillment. The top 5 features are:
But-Present, Post-Context-Connotation-Disagree,
Post-Context-Violated-Expectation, Desire-Verb, Is-First-Person.  We also
tested different feature sets separately. We describe our experiment
results below.

The results of the feature ablation experiments using LR model are shown in
Table~\ref{tab:feature}. The ALL feature set includes all the features
described in Sec.~\ref{sub:features} (without BOW). We obtained high precision and F-measure using the Discourse features. We also experimented
with our top feature from the InfoGain analysis, \textit{But-Present},
which surprisingly achieves a high F-measure, compared to using ALL
and Discourse feature sets. The last row of Table~\ref{tab:feature}
shows the results of using ALL features excluding
\textit{But-Present}. This indicates that features motivated by
narrative structure are primarily driving improvement.  In previous
work ~\newcite{Chaturvedietal16} show that a model representing
narrative structure could beat the BOW baseline, but they performed no
systematic feature ablation.  Our results suggest that ultimately, the
presence of \textit{``but"} is likely a central driver for their
improvements as well.

\begin{table}[t]
\footnotesize
\centering
\begin{tabularx}{\columnwidth}{p{1.25in} | c c c }
\toprule
\bf Features &  \bf Precision & \bf Recall & \bf F1\\
\midrule
ALL & 0.64 & 0.64 & 0.64 \\
Discourse & 0.66 &  0.64 & 0.65 \\
But-Present & \bf 0.72 & \bf 0.64 & \bf 0.68 \\
ALL w/o But-Present &  0.58 & 0.58 & 0.58 \\
\bottomrule
\end{tabularx}
\caption{\label{tab:feature} Results of Logistic Regression classifier with different feature sets on Simple-DesireDB}
\end{table}

\subsection{Comparison to Previous Work}  
\label{sub:comparison}
We directly compare our methods and features to the most relevant previous work~\cite{Chaturvedietal16}. They applied their models on two datasets and reported the results for the Fulfilled class. We present the same metrics in Table~\ref{tab:chaturvedi-compare}, using our best model \textbf{Skip-Thought} (SkipTh). We also present results of our LR model with our Discourse features, \textbf{Discourse-LR}, trained and tested on their corpora to compare to their features. The first three rows show the results from \newcite{Chaturvedietal16} for comparison. As described in Sec.~\ref{related}, they used BOW as baseline, LSNM is their best model, and Unstruct-LR is their unstructured model that uses all of their features with LR.

On both corpora, \textbf{Discourse-LR} outperforms Unstruct-LR, showing that the Discourse features are stronger indicators of the desire fulfillment status when used with LR classifier. 
In addition, on SimpleWiki, LR-Discourse outperforms their structured model, LSNM (0.46 vs. 0.27 on F-1).

\begin{table}[t]
\footnotesize
\centering
\begin{tabularx}{\columnwidth}{p{0.4in}| p{0.75in}|c c c}
\toprule
\bf Dataset & \bf Method & \bf Precision & \bf Recall & \bf F1 \\
\midrule
MCTest & BOW & 0.41 & 0.50  &  0.45  \\
 & Unstruct-LR & 0.71 & 0.63 &  0.67 \\
 & LSNM & 0.70 & 0.84 &  0.74  \\
& \bf Discourse-LR  & \bf 0.63 & \bf 0.83  & \bf 0.71 \\
& \bf SkipTh-BOW & \bf 0.72 & \bf 0.68 & \bf 0.70 \\
& \bf SkipTh-ALL & \bf 0.70 & \bf 0.84 & \bf 0.76 \\
\midrule
Simple & BOW & 0.28  & 0.20 & 0.23   \\
Wiki & Unstruct-LR & 0.50 & 0.09 & 0.15 \\
 &  LSNM & 0.38 & 0.21 &  0.27 \\
 & \bf Discourse-LR  & \bf 0.32 & \bf 0.82 & \bf 0.46 \\
 & \bf SkipTh-BOW & \bf 0.71 & \bf 0.26 & \bf 0.38 \\
 & \bf SkipTh-ALL & \bf 0.33 & \bf 0.16 & \bf 0.22 \\
\bottomrule
\end{tabularx}
\caption{\label{tab:chaturvedi-compare} Previous work 
and {\bf our results} for the Fulfilled class, on MCTest and SimpleWiki.}
\end{table}

\section{Conclusion and Future Work}
\label{conclusion}

We created a novel dataset, DesireDB, for studying the expression of
desires and their fulfillment in narrative discourse. We show that
contextual features help with classification, and that both prior and
post context are useful. Finally, we show that exploiting narrative 
structure is helpful, both directly in terms of the utility of discourse
relation features and indirectly via the superior performance of a Skip-Thought
LSTM model. 

In future work, we plan
to explore richer features and models for semantic and discourse-based features,
as well as the utility of more narratively-aware features. For instance, the
sentiment flow features roughly track the notion that the arc of a
narrative may implicitly reveal resolution of a goal via changes in affect states.
We hope to examine whether there are other similar rough-grained measures of
change over the entire narrative that can improve the results. 

DesireDB contains annotator-labeled spans for evidence for the
annotator's conclusions. While we have not used this labeling, we
plan to use it in future work. 
Finally, we hope to turn to automatically detecting instances of desire
expressions that give rise to the kind of goal-oriented narratives
DesireDB contains. Here we have used high-precision search patterns
but our annotations  show that such patterns still admitted 134
hypothetical desires (e.g., `If I had wanted to buy a book'). It would
appear that distinguishing hypothetical vs. real desires itself could
be an interesting problem.

%
%
%
%
%
%
%

\section*{Acknowledgments}
This research was supported by Nuance Foundation Grant SC-14-74, NSF Grant IIS-1302668-002 and IIS-1321102.

\bibliography{nl}
\bibliographystyle{acl_natbib}



\end{document}